\documentclass[letterpaper, 10 pt, conference]{ieeeconf} 
\IEEEoverridecommandlockouts

\usepackage{amsmath,amssymb,amsfonts}
\usepackage{algorithmic}
\usepackage{graphicx}
\usepackage{textcomp}
\usepackage{xcolor}
\usepackage{soul}

\author{Martin Bikandi$^{1,2}$, Gorka Velez$^{1}$, Naiara Aginako$^{2}$ and Itziar Irigoien$^{2}$
\thanks{$^{1}$ 
        Vicomtech Foundation, Basque Research and Technology Alliance (BRTA), Mikeletegi Pasealekua, 57, 20009 Donostia-San Sebastián, Spain
        \tt\small \{mbikandi,gvelez\}@vicomtech.org}%
\thanks{$^{2}$ 
        UPV/EHU, Faculty of Informatics, Manuel Lardizabal pasealekua, 1, 20018 Donostia-San Sebastián, Spain
        \tt\small \{naiara.aginako, itziar.irigoien\}@ehu.eus}%
}

\title{\LARGE \bf Synthetic outlier generation for anomaly detection in autonomous driving}

\begin{document}

\maketitle

\begin{abstract}
Anomaly detection, or outlier detection, is a crucial task in various domains to identify instances that significantly deviate from established patterns or the majority of data. In the context of autonomous driving, the identification of anomalies is particularly important to prevent safety-critical incidents, as deep learning models often exhibit overconfidence in anomalous or outlier samples. In this study, we explore different strategies for training an image semantic segmentation model with an anomaly detection module. By introducing modifications to the training stage of the state-of-the-art DenseHybrid model, we achieve significant performance improvements in anomaly detection. Moreover, we propose a simplified detector that achieves comparable results to our modified DenseHybrid approach, while also surpassing the performance of the original DenseHybrid model. These findings demonstrate the efficacy of our proposed strategies for enhancing anomaly detection in the context of autonomous driving.

\end{abstract}


\begin{center}
\end{center}

\section{Introduction}

Deep learning models use huge amounts of data to learn specific tasks. Recently, these models are able to perform even better than humans, but only under certain conditions: inference of deep learning models on data that differs too much from the training data often leads to incorrect or overconfident predictions. 
Being able to detect outlier or anomalous data can be useful, for example, for error detection in biomedical images or production lines and detecting unexpected behaviour in video surveillance cameras \cite{survey}. 

In autonomous driving, in particular, anomaly detection is crucial since an incorrect prediction may lead to an accident. In case of a road anomaly such as a fallen tree or light pole, a hole in the road, or an uncommon vehicle in circulation, detecting it firsthand and requiring human intervention may avoid such accidents. We want an accurate and fast way to detect the uncertainties on the road that produces low amounts of false positives for fewer human interventions \cite{surveyautonomous}.

Allowing a model to work properly in the open world, where the data is uncertain, as opposed to a closed world with a set number of properly labelled classes and environments, is a challenging problem. Novel classes arise everywhere, and handling these rare occurrences is only possible if detection is possible first. One of the reasons autonomous vehicles are not commonplace and only work in certain locations and/or conditions is that corner cases are many, unknown, and difficult to model. 

In this article, we explore the current research of anomaly detection based on semantic segmentation and propose simple changes to the generation of training examples with synthetic outliers, which produce better results. We test these changes to the synthetic outlier generation by training the model DenseHybrid \cite{dh} with some modifications to the cost function that give slightly better results. We also propose a simple approach to turn segmentation models into anomaly detectors, which we call Simplified Detector (SD).

We extensively test the resulting models on a subset of the database SHIFT \cite{shift}, which we filtered to avoid pixels labelled as a pedestrian during training, effectively turning this class into an anomaly. Filtering out the pedestrian class is suitable to test anomaly detection methods since the variability of this class is high in colours and shapes, so detecting pedestrians via an anomaly detection method can reflect real open-world problems. Lastly, we test the proposed models on the public dataset StreetHazards to compare the model to the current research.

The rest of the paper is organized as follows: In section \ref{related_works}, we overview the approaches for anomaly detection in autonomous driving. In section \ref{proposed_method}, we explain the proposed changes. In section \ref{experimental_setup}, we overview the datasets, metrics and baselines used in the experiments and in section \ref{experimental_results}, the obtained results are shown. In section \ref{conclusions}, the conclusions are presented.



\section{Related Work}
\label{related_works}

Anomaly detection can be split into two broad categories: supervised and unsupervised anomaly detection. Due to the difficulty of finding labelled anomalous data in most scenarios, unsupervised anomaly detection is the default choice. Many recent approaches use datasets disjoint from the training data but of the same type (e.g. image) and use the observations from those datasets as anomalies. This data is inserted in various ways on the training data and treated as an anomaly during training.

Apart from this distinction based on the available data, anomaly detection techniques can be split based on the approach. In the case of anomaly detection in images for autonomous driving, approaches are mainly the following: \textit{Reconstruction, Generative, Confidence scores and Feature modelling}.

\textit{Reconstruction} methods use encoder-decoder networks to resynthesize an input image. The anomalies are detected by means of the reconstruction error \cite{synboost}. \textit{Generative} approaches use generative networks to model uncertainty during training or inference \cite{nflow}. \textit{Confidence score} based methods form a baseline for anomaly detection, since they use the uncertainties in the prediction of the neural network as a means to measure outlierness \cite{dh, sml, sh, dml}. Instead of detecting anomalies in the actual image, \textit{feature modelling} based methods try to find the anomalies in a representation space. The representation is constructed by hand or learned \cite{moose, cdpn}.

To generate synthetic outliers, one can use the approach from \cite{oe}, Outlier Exposure, using an external dataset and pasting images randomly sampled from this dataset on the labelled inputs, labelling the pasted pixels as anomalous. Another approach is to use a generative model to produce these anomalous pixels, using for instance, a Diffusion model, a generative adversarial network (GAN) or a normalizing flows model \cite{nflow}.

Most recent works on anomaly detection for autonomous driving use semantic segmentation models with various approaches to detect outliers. Some of the methods are computationally costly since the uncertainty is calculated with several passes of the network by slightly modifying the input image. Many methods use negative auxiliary datasets to produce synthetic outliers or start from a trained model and fine-tune on small datasets with labelled outliers.

Anomaly detectors based on semantic segmentators are a good choice to detect anomalies since the models are trained to produce good segmentation results while making good anomaly predictions. This may result in a good and efficient multitask model: a single pass from the network produces the semantic segmentation and anomaly scores for each pixel, which can be used separately for other tasks \cite{survey, surveyautonomous}.

\section{Proposed Method}
\label{proposed_method}

We build upon the DenseHybrid model \cite{dh}, which itself builds upon semantic segmentation models to include anomaly detection capability. It is a confidence score-based method which is trained on images where synthetic outliers are introduced randomly. We introduce changes only to the training of the model. First, by slightly modifying the cost function used to train it, and second, by producing synthetic outliers differently.



The semantic segmentation model we use during testing is based on the DeepLabV3+ (DLV3+) \cite{dlv3p} architecture. The backbone (feature extractor) of the model can be any convolutional neural network (CNN), but we use the Wide-ResNet-38 network \cite{wrn}. 

\subsection{Changes to DenseHybrid}

\begin{figure}[ht]
    \centering
    \includegraphics[width=6cm]{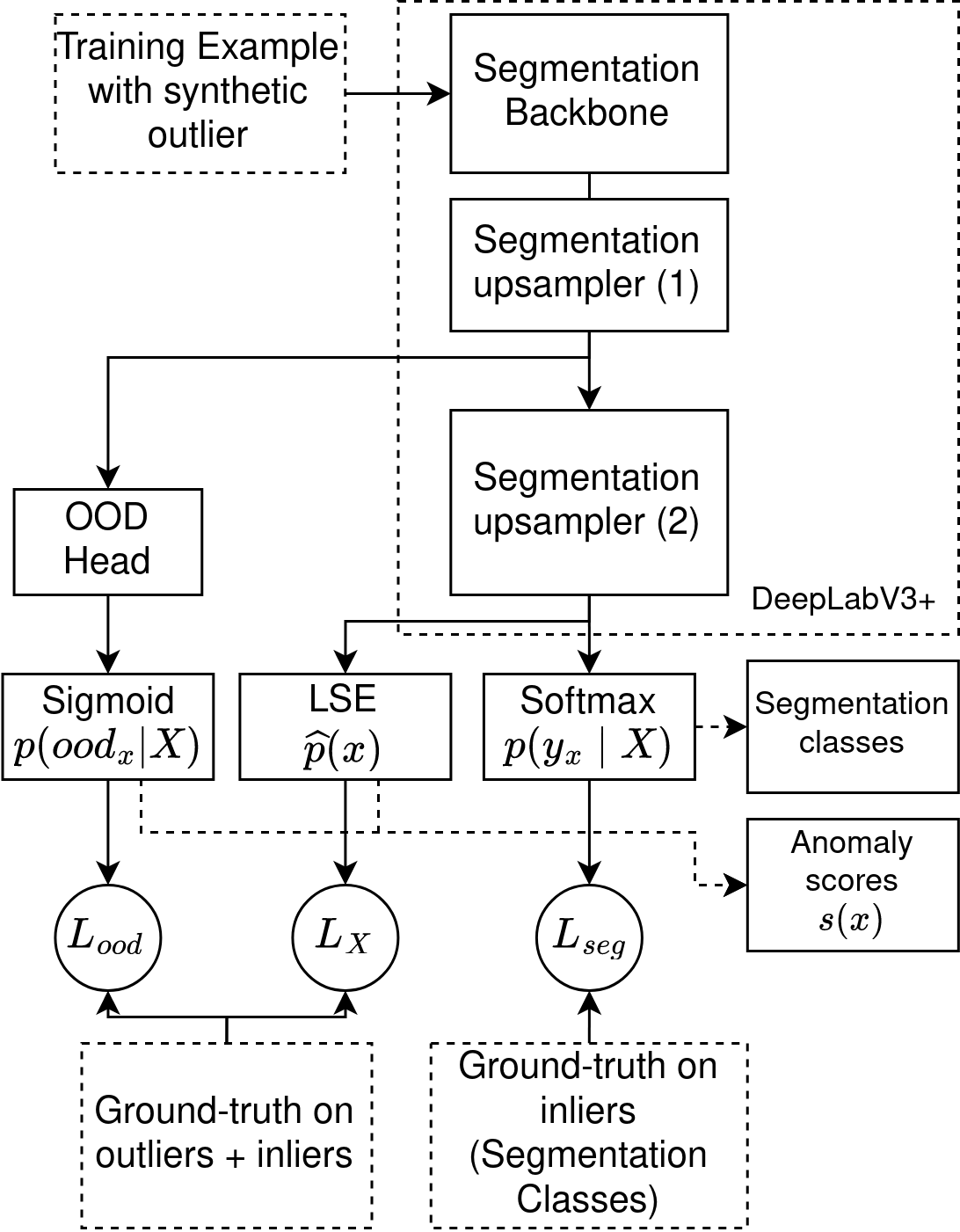}
    \caption{DenseHybrid architecture. The model is trained with mixed content training examples, which affect each part of the cost function differently. We propose changes to the generation of training examples with synthetic outliers and also changing $L_{X}$.}
    \label{fig:dh}
\end{figure}

The architecture summary of DenseHybrid is presented in the diagram of Figure \ref{fig:dh}. The segmentation model is complemented with an out-of-distribution (OOD) head starting from the first step of the segmentation upsampler. This OOD head produces probabilities $p(ood_x|X)$, that is, the probability of pixel $x$ being an outlier given the image $X$. 


Additionally, the model produces anomaly scores based on a reinterpretation of the segmentation logits to recover the unnormalized data likelihood. The model then combines the output from the OOD head and the likelihood in a hybrid score, hence the name.

The model is trained on mixed content images containing synthetic outliers from a negative database. In the original paper, the model is explicitly trained to: 

\begin{itemize}
    \item Improve the segmentation performance on inliers
    \item Improve the classification performance of the OOD head on synthetic outliers vs inlier
    \item Minimize the log-likelihood of synthetic outliers
\end{itemize}

The model is also implicitly trained to maximize the log-likelihood of inliers. This is a result of optimizing the segmentation network.

The change to the cost function we propose is to also explicitly maximize the log-likelihood of inliers. Since the minimization of the log-likelihoods is explicit, we expect that explicit maximization of the likelihood of inliers will improve the detection performance of the model.

Let $(x, y_x)$ be a pixel of a sample image, with its corresponding class, $p(x)$ its likelihood and $\hat{p}(x)$ its unnormalized likelihood. If a pixel is in-distribution, we denote it by $x\sim D_{in}$ or $(x, y_x)\sim D_{in}$, otherwise it is $x\sim D_{out}$. The model produces probabilities $p(y_x|X)$, i.e. probability of pixel $x$ being of class $y$, and $p(ood_x|X)$, explained previously.

The idea behind DenseHybrid is to interpret the exponentiated outputs of the model as the joint unnormalized likelihood:

$$\hat{p}(x, y) = exp(s^y_x)$$



Here $s^y_x$ corresponds to the logit of the classifier for class $y$ at pixel $x$. Due to this interpretation, the unnormalized likelihood can be extracted from the model's segmentation output via a sum-exp operation over the logits:

$$\hat{p}(x) = \sum_{y} exp(s^y_x) = \sum_{y} \hat{p}(x,y)$$

These are the parts of the cost function as explained in the original paper \cite{dh}:

\begin{align*}
    L_X &=\mathbb{E}_{x\sim D_{in}}[-\ln (\hat{p}(x))] - \mathbb{E}_{x\sim D_{out}}[-\ln (\hat{p}(x))] \\
    &=L_{X_{in}} + L_{X_{out}} \\
    L_{seg} &= \mathbb{E}_{(x,y)\sim D_{in}}[-\ln (p(y_x|X))] \\
    L_{ood} &= \mathbb{E}_{x\sim D_{in}}[-\ln (1-p(ood_x|X))] \\
    &+\mathbb{E}_{x\sim D_{out}}[-\ln (p(ood_x|X))]
\end{align*}

In \cite{dh}, the resulting loss is the following sum, weighted with a loss modulation parameter $\beta$:

$$L_{DH} = \beta L_{X_{out}} + 10\beta L_{ood} + L_{seg}$$

The loss $L_{X_{in}}$ is not in the sum since this cost is implicitly optimized by $L_{seg}$, as explained in the original paper \cite{dh}.

We propose using $L_{X_{in}}$ during the training, weighted by its own hyperparameter, to have more control over the training. Thus, the resulting new cost function is:

$$L_{DH2} = \beta_1 L_{X_{out}} + \beta_2 L_{X_{in}} + \beta_3 L_{ood} + L_{seg}$$


The original DenseHybrid cost can be recovered by selecting the $\beta$ hyperparameters as $(\beta_1, \beta_2, \beta_3) = (\beta, 0, 10 \cdot\beta)$.

DenseHybrid produces anomaly scores based on the following formula:

$$s(x) = \ln\left(\frac{p(ood_x|X)}{\hat{p}(x)} \right) = \ln (p(ood_x|X)) - \ln(\hat{p}(x))$$

During testing, we test the original anomaly score, based on the log ratio between the OOD head probabilities and the unnormalized log-likelihood, together with other baselines.

\subsection{Simplified Detector}
\label{simplified}

We propose to use the reinterpretation of logits done in \cite{dh} to train a model, by maximizing the likelihood of inliers and minimizing the likelihood of outliers. This modification could be applied to any segmentation network architecture, and the anomaly scores produced by the model would be inferred only from the segmentation logits. This score could be the unnormalized log-likelihood or simply the negative maximum logit.

We propose using the following cost, where each piece comes from the previous section:

$$L_{SD} = \beta L_{X_{out}} + \beta L_{X_{in}} + L_{seg}$$

We simply remove the part of the cost function affecting the OOD head. This is in fact a simple change from the modified DenseHybrid model. Taking the $\beta$ hyperparameters as $(\beta_1, \beta_2, \beta_3) = (\beta, \beta, 0)$ would have the effect of gradients not propagating through the OOD head, and thus only affecting the segmentation model.

\subsection{Mixed content image generation}

DenseHybrid depends on synthetic outliers to be properly trained. How these mixed content images are crafted is crucial for properly training the model.

Generating synthetic outliers simply by pasting an image from the negative dataset on top of the training example can be problematic: synthetic outliers pasted directly on the image may result in a rectangular negative patch on uninteresting regions of the image. Randomly pasting samples from the negative dataset may result in inlier objects being treated as outliers during training too, if those inlier classes are not filtered first. Pasted regions may also be too apparent and easily classified by the model because they have too different colours compared to the original training example, or because of sharp edges around the anomaly.


\subsubsection*{Filtering the negative dataset}

Pasting cars, road pieces, or classes included in the training is counter-productive since pasted regions are treated as outliers during training. We filter out the negatives to exclude all classes included in the training dataset, effectively solving this problem. We filter classes such as car, pedestrian, road, etc, and paste outliers from a smaller but more outlier dataset.

\subsubsection*{Pasting below horizon}
To avoid pasting outliers in the region of the image corresponding to the sky or to background buildings, and since all the images of the datasets have similar camera orientations, we trace a horizon line in the images and only paste outliers below that line. This way, outliers are only pasted on regions where outliers are more likely to appear and are more critical for a vehicle. That is, on the ground. 


In order to blend the pasted anomaly into the original training example, we try several techniques. Our goal is to produce images in which the pasted regions are properly blended into the environment, similar to what is done in the dataset Fishyscapes Static \cite{fs}.

\subsubsection*{Alpha blending}

First of all, we implemented alpha blending to paste outliers in the training example. The blending function assigns certain weights to each pixel of the input images, adds them together and outputs the result into a new pixel. In that way, the pasted outliers give an impression of transparency.

\subsubsection*{Synthetic outlier shape}

In the original DenseHybrid paper, the pasted outliers have a rectangular shape with 50\% probability. The rest of the times, the pasted outlier only includes a couple of classes from the labels of that negative image. To produce outliers with more varied shapes we propose avoiding rectangular pasted outliers.

\subsubsection*{Histogram matching}

In order to make the colours of the pasted pixels similar to the ones in the background image, we use histogram matching to adapt the colours of the pasted pixels. This procedure results in an image with more uniform colours, and not-so-sharp changes from background to pasted pixels \cite{histogram}.

\subsubsection*{Gaussian blurring}

Edges of the pasted outlier can be very sharp, so we try to smooth the pasting by applying Gaussian blurring to the alpha mask of the negative patch.

\section{Experimental Setup}
\label{experimental_setup}

\subsection{Datasets}

For the experiments, we use the following datasets:

\subsubsection*{SHIFT}

The SHIFT dataset \cite{shift} is a synthetic dataset for synthetic-real domain shift. Due to the large amount of training data available, filtering this database is possible: training images without pixels classified as pedestrian form a subset of the database with 23K images. We call this database SHIFT Nopedestrian.

For testing, we filter out the database the other way, considering only images with a sufficient amount of Pedestrian pixels from the validation subset. We consider a low and high threshold of 10K and 15K pedestrian pixels (in 1MP images), we call this dataset SHIFT Pedestrian, with a total of 1070 images.

\subsubsection*{StreetHazards}

StreetHazards is a public anomaly detection dataset for testing presented in \cite{sh} It is a small dataset of 1500 images with anomalies of various sizes, properly labeled. It includes a training set of $\sim$5K images, which we use to train the models.

\subsubsection*{ADE20K}

The ADE20K dataset is a segmentation dataset provided by the MIT \cite{ade}. We use this dataset as the negative dataset for synthetic outlier generation.

Similar to filtering the SHIFT dataset, we filter the ADE Challenge database to exclude classes included in the training. The resulting filtered dataset is quite small compared to the original 20K images, for a total of 487 negative images.

\subsection{Metrics}

Outlier detection is a binary classification task. The approach of outlier detection is to give each pixel in the image a score based on the model to measure the outlierness of said pixel. The higher the score, the more likely the pixel is an outlier. 

The metrics used for anomaly detection are Average Precision (similar to area under the precision-recall curve, AUPRC) and False Positive Rate at 95\% True Positive Rate (FPR$_{95}$ or FPR@95). We also measure the area under the receiving operating characteristic curve (AUROC) which is used to measure binary classification performance. To measure the segmentation performance, we use the mean intersection over union (mIoU), also known as Jaccard index. 


\subsection{Baselines}

Anomaly detection is a complicated task with many possible approaches. There are some baselines that can be used to compare the results of more complex models.

\subsubsection*{MSP}

Max softmax probability (MSP) is a baseline for outlier detection, which relies on the confidence of classification to produce an anomaly score \cite{sh}.

\subsubsection*{Max Logits}

Max logits is the refined version of the max softmax probability baseline \cite{sh}, which uses the pre-activations of the last layer of the classifier as the anomaly scores.



\section{Experimental Results}
\label{experimental_results}
We experiment by training on the SHIFT Nopedestrian dataset and evaluating the performance for each proposed modification in the training. In every experiment, we use the same $\beta$ hyperparameters, $(0.01, 0.005, 0.2)$ for the modified cost and $(0.01,0,0.1)$ for the original DenseHybrid model. The beta hyperparameters for the modified cost are chosen to be similar to the ones used in the original DenseHybrid model. Since the likelihood maximization is implicitly optimized, we propose half the beta value to weight this part of the optimization (0.005). For the Simplified detector, the $\beta$ hyperparameter is set to 0.01. 

\subsection{Evaluation of Changes to DenseHybrid Training}

We refer to the original model from \cite{dh} as DH and the model trained with the modified cost function as DH2. Just changing the cost function slightly improves the results with respect to the original DH. Therefore, we test the rest of the changes over DH2. The results of the proposed changes (modifying the cost function, considering the horizon while pasting, etc) are summarized in Table \ref{tab:results_changes}.

\begin{table}[ht]
    \centering
    \begin{tabular}{l|ccc|c}
        Model & AP$\uparrow$ & FPR$_{95}$$\downarrow$ & AUROC$\uparrow$ & mIoU\\
        \hline
        DH & 18,34 & 24,14 & 93,99 & 72,99\\
        DH2 & \textbf{19,2} & \textbf{21,09} & \textbf{94,77} & \textbf{73,04}\\
        \hline
        DH2 + Below Horizon (1) & 23,9 & 19,35 & 95,46 & 73,13 \\
        DH2 + Blending (2) & 18,72 & 20,09 & 94,86 & 73,12 \\
        DH2 + Filter Negatives (3) & 21,99 & 18,87 & 95,44 & \textbf{73,42}\\
        DH2 + No Rectangles (4) & 20,7 & 19,87 & 95,07 & 73,04\\
        DH2 + Blur (5) & 17,66 & 17,65 & 95,04 & 72,75\\
        DH2 + Histogram M. (6) & 19,11 & 14,54 & 95,81 & 72,32\\
        DH2 + (1,2,3,4,5,6) & \textbf{33,97}	 &\textbf{ 7,72} &	\textbf{97,73} & 72,82
    \end{tabular}
    \caption{Anomaly detection performance on SHIFT Pedestrian for different synthetic outlier generation methods, best results in bold}
    \label{tab:results_changes}
\end{table}

The mIoU over the inlier classes of the test images is quite similar across all the models, being the change between the original DH model and the rest of less than $1\%$ mIoU.


We can see how the biggest improvement in AP from an individual change comes from pasting the synthetic negatives below the horizon. This is probably a consequence of the test dataset containing only anomalies in the lower part of the image since anomalies are pedestrians. Pasting synthetic outliers where anomalies are not usually found reduces the overall anomaly detection performance of the model.

Applying the blending operation, blurring, and histogram matching are additions to the training that slightly lowers the performance in AP while improving the FPR. The biggest improvement in FPR comes from histogram matching of the pasted negative.


The best model was the DH2 model with all changes: Pasting under horizon + Alpha blending + No rectangles as outliers + Filtering the negatives dataset + Gaussian Blurring + Histogram Matching. The resulting model outperforms the original DH model by $15.63\%$ AP, and the original DH2 model by $14.77\%$ AP. This is a significant increase in performance by simple changes in the training process. In the FPR$_{95}$ metric, the best model outperforms the DH model by $16.42\%$ and the DH2 model by $13,37\%$.



\subsection{Evaluation of Baselines and Simplified Detector}

To evaluate the baselines and the simplified detector, we use the following models:

\begin{enumerate}
    \item Base DLV3+ model
    \item DH2 + (1,2,3,4,5,6). We call this model DH2 ALL for short.
    \item Simplified Detector + (1,2,3,4,5,6). We call this model SD ALL for short.
\end{enumerate}

We evaluate the baselines on the semantic segmentation model from which all the rest were fine-tuned by using the three baselines proposed, in addition to the scoring function based on the negative unnormalized log-likelihood $\hat{p}(x)$ as introduced in section \ref{proposed_method}. Table \ref{tab:results_filter} summarizes the results obtained from the different baselines, compared to the baselines applied to the underlying segmentation model.


\begin{table}[ht]
    \centering
    \begin{tabular}{l|ccc}
        Model & AP$\uparrow$ & FPR$_{95}$$\downarrow$ & AUROC$\uparrow$ \\
        \hline
        DLV3+ MSP & 7,1 & \textbf{37,35} & 87,28\\
        DLV3+ ML & 11,58 & 42,8 & 89,29\\
        DLV3+ $\hat{p}(x)$ & \textbf{12,49} & 42,89 & \textbf{89,5}\\
        \hline
        DH2 ALL + Original score& 33,97 & 7,72 & 97,73 \\
        DH2 ALL + MSP & 26,38	&21,32&	94,18\\
        DH2 ALL + ML & \textbf{34,87}	&7,75&	97,73\\
        DH2 ALL + $\hat{p}(x)$ & 34,38	&\textbf{7,31}&	\textbf{97,74}\\
        \hline
        SD ALL + MSP & 15,05	&22,29&	92,78\\
        SD ALL + ML & \textbf{31,5}	&\textbf{9,2}&	\textbf{97,4}\\
        SD ALL + $\hat{p}(x)$ & 30,36	&9,19&	97,38\\
    \end{tabular}
    \caption{Anomaly detection performance of the baselines, DH2 ALL and the Simplified detector, best results in bold}
    \label{tab:results_filter}
\end{table}



We were surprised to find that the hybrid anomaly score is in fact worse in average precision than simply considering the max logits or negative unnormalized likelihood as scores. Due to simplicity, max logits seems to be the most reasonable scoring function to consider.


The cost function that encourages maximization of likelihood on inliers and minimizing likelihood on outliers is, in fact, maximizing logits on inliers and minimizing logits on outliers, so the result could be expected.


Table \ref{tab:results_filter} shows that the Simplified Detector, while not matching the performance of DH2 ALL, is close to it in every metric. Comparing to the numbers of Table \ref{tab:results_changes}, we see that the simplified detector still outperforms the basic DH model by a big margin (and also the basic DH2 model), with 13,16\% AP improvement and 14,94\% lower FPR$_{95}$. Concerning the DH2 ALL model with max logits, the simplified detector falls behind in AP by about 3\% 

\subsection{Evaluation on StreetHazards}

\begin{figure}[ht]
    \centering\includegraphics[width=\columnwidth]{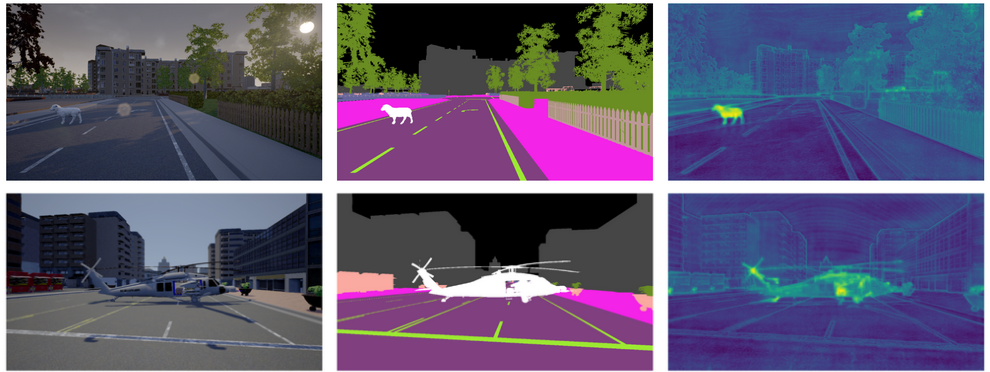}
    \caption{Inference of the DH2 NHNB model for StreetHazards, anomalies are painted white in the ground truth (centre image). The RGB image and the anomaly scores are on the left and right, respectively.}
    \label{fig:inference_example}
\end{figure}

In order to compare the proposed changes to the state-of-the-art (SOTA), we train and test the model on the StreetHazards (SH) dataset. Similar to the evaluation on SHIFT, we train the following models on StreetHazards:

\begin{enumerate}
    \item Base DLV3+ model
    \item DenseHybrid original model
    \item DH2 + (3,4,5,6), i.e. the model with all the changes except pasting under the horizon and pasting with alpha blending, we call this model DH2 NHNB (No Horizon, No Blend) for simplicity
    \item Simplified Detector NHNB (SD + (3,4,5,6))
\end{enumerate}

The results are summarized in Table \ref{tab:results_sh}. We selected DH2 NHNB instead of DH2 ALL since the anomalies in the StreetHazards dataset can be found in the upper parts of the image, and through experimentation we found that avoiding this step slightly improved results. We also found that Blending was not improving results in the case of StreetHazards.

\begin{table}[ht]
    \centering
    \begin{tabular}{l|ccc}
        Model & AP$\uparrow$ & FPR$_{95}$$\downarrow$ & AUROC$\uparrow$ \\
        \hline
        DLV3+ MSP& 8,7 & \textbf{24,45} & 91,45\\
        DLV3+ ML& \textbf{11,29} & 36,19 & 91,54 \\
        DLV3+ $\hat{p}(x)$ & 11,27 & 36,28 &\textbf{ 91,58}\\
        \hline
        DH & 31,4 & 13,84 & 96,37\\
        DH2 NHNB + Original score & \textbf{37,37}	&\textbf{11,4}&	\textbf{97,31}\\
        DH2 NHNB + MSP & 21,25	&17,69&	94,53\\
        DH2 NHNB + ML & \underline{36,99}	&12,69&	97,03\\
        DH2 NHNB + $\hat{p}(x)$ & 36,73&\underline{12,36}&	\underline{97,12}\\
        SD NHNB + MSP & 20,6	&17,87&	94,64\\
        SD NHNB + ML & 34,65	&13,93&	96,64\\
        SD NHNB + $\hat{p}(x)$ & 33,86	&13,74&	96,65\\
    \end{tabular}
    \caption{Anomaly detection performance on StreetHazards with DH NHNB and SD NHNB compared to the baseline, best results in bold, second best underlined}
    \label{tab:results_sh}
\end{table}

The proposed changes are able to improve the anomaly detection over the original DenseHybrid method by up to $5,97\%$ AP and $2,42\%$ FPR$_{95}$. The simplified detector based on max logits scoring function is still able to outperform the original DenseHybrid approach by a small margin in both metrics but still falls behind when compared to the DH2 NHNB model. Figure \ref{fig:inference_example} shows inference examples from our DH2 NHNB model.

\subsection{Comparison to the SOTA}

Lastly, we compare the best results obtained on the StreetHazards dataset to several methods of the literature. The results are summarized in Table \ref{tab:results_sh_sota}.

\begin{table}[ht]
    \centering
    \begin{tabular}{l|c|ccc}
        Model & OE & AP$\uparrow$ & FPR$_{95}$$\downarrow$ & AUROC$\uparrow$ \\
        \hline
        DH (LDN-121)\cite{dh} & \checkmark & 30,2 & 13,0 & 95,6\\
        NFlowJS \cite{nflow} & & 28.4 & 14.9 & 95.7\\
        MOoSe (ML) \cite{moose} &  & 15.22 & 17.55 & - \\
        DML \cite{dml} &  & 14,7 & 17,3 & 93,7 \\
        cDPN \cite{cdpn} & & \textbf{46,2} & 14,9 & - \\
        cosMe \cite{cosme} & & 19,7 & 15,5 &  94,6\\ 
        \hline
        DH (DLV3+) & \checkmark& 31,4 & 13,84 & 96,37\\
        DH2 NHNB &  \checkmark& \underline{37,37}	&\textbf{11,4}&	\textbf{97,31}\\
        DH2 NHNB + ML & \checkmark& 36,99	&\underline{12,69}&	\underline{97,03}\\
        SD NHNB + ML &  \checkmark&34,65	&13,93	&96,64\\
    \end{tabular}
    \caption{Anomaly detection performance on StreetHazards compared to other methods in the research, best results in bold, second best underlined. The OE column indicates if the model was trained with additional negative training data}
    \label{tab:results_sh_sota}
\end{table}

Our method outperforms every other in AP, except for cDPN \cite{cdpn}. Note that \cite{cdpn} is only published in arXiv and is not peer-reviewed yet. Compared to the original DenseHybrid model, we get more than $7\%$ AP increase while slightly reducing the FPR$_{95}$. The proposed simplified detector also outperforms the original DenseHybrid model, improving the AP by more than 4\% in exchange for a slightly worse FPR$_{95}$.

Comparison between models is not completely fair since not all methods are based on the same segmentation model. Still, our method DH2 NHNB achieves SOTA metrics on the dataset StreetHazards, obtaining the best results among the peer-reviewed works. 

\section{Conclusions and Future Work}
\label{conclusions}

We conclude from the experiments that filtering out the database containing the negative outliers, and pasting outliers where the anomalies are expected does significantly improve the anomaly detection performance of the model. Pasting anomalies with the shape of objects in the negative database, adjusting the colour of the pasted pixels, blurring the edges, and pasting with alpha blending also help, even if the effects are small on their own. Combining all the proposed training changes, including the change to the cost function, results in a big improvement over the original DenseHybrid model.

We found that the resulting model performed unexpectedly well when used with the simple Max Logits scoring function. For this reason, we proposed the Simplified Detector. This method, while not achieving the best results, outperforms the base model while requiring less computation, since the OOD head is avoided and can be used on any segmentation model.


In conclusion, the preprocessing of the negative dataset, and more elaborate pasting of synthetic outliers on training samples improves performance compared to the original approach. With these simple changes, we were able to improve the AP over the original model by more than 7\% and achieved the best FPR$_{95}$ when compared to other anomaly detection methods of the SOTA.

Future work includes applying filtering operations to the model output and improving the performance of the simplified detector without the OOD Head. This could be done via slight additional changes to the cost function. Training the models from scratch instead of fine-tuning a pretrained classifier might lead to improved results too.


\section*{ACKNOWLEDGMENT}
This work is part of the 5GMETA project. This project has received funding from the European Union's Horizon 2020 research and innovation program under grant agreement No 957360. Content reflects only the authors' view and European Commission is not responsible for any use that may be made of the information it contains.

\bibliographystyle{IEEEtran}
\bibliography{ref}

\end{document}